\documentclass{article} %
\usepackage{iclr2027_conference,times}

\usepackage{amsmath,amsfonts,bm}

\def\eqref#1{equation~\ref{#1}}

\def\1{\bm{1}}

\def\ie{{\textit{i.e.}}}
\def\etc{{\textit{etc}}}

\DeclareMathAlphabet{\mathsfit}{\encodingdefault}{\sfdefault}{m}{sl}
\SetMathAlphabet{\mathsfit}{bold}{\encodingdefault}{\sfdefault}{bx}{n}

\DeclareMathOperator*{\argmax}{arg\,max}
\DeclareMathOperator*{\argmin}{arg\,min}

\usepackage{hyperref}
\usepackage{url}
\usepackage{graphicx}
\usepackage{booktabs}
\usepackage{multirow}
\usepackage{xcolor}
\usepackage[normalem]{ulem}
\usepackage{float}
\usepackage{algorithm}
\usepackage{algpseudocode}
\usepackage{placeins}
\usepackage[capitalize]{cleveref} %
\crefname{section}{Sec.}{Secs.}
\Crefname{section}{Section}{Sections}

\newcommand{\best}[1]{\textbf{#1}}
\newcommand{\second}[1]{\underline{#1}}

\definecolor{keycolor}{RGB}{0,45,120}

\graphicspath{{src/figure/}{src/appendix/figure/}}

\title{Point2Part: \\ Unified 3D Partitioning from Point Prompts}

\author{
\bf Hao-Tang Tsui \qquad Yu-Rou Tuan \qquad Xiaoxuan Ma \\[1pt]
\bf Nicol\'as Ugrinovic \qquad Takaaki Shiratori \qquad Kris Kitani \\[4pt]
Carnegie Mellon University \\[3pt]
\url{https://henrytsui000.github.io/Point2Part}
}

\iclrfinalcopy
\begin{document}

\maketitle
\vspace{-3em}
\lhead{Point2Part: Unified 3D Partitioning from Point Prompts}

\begin{figure}[H]
\centering
\includegraphics[width=\linewidth]{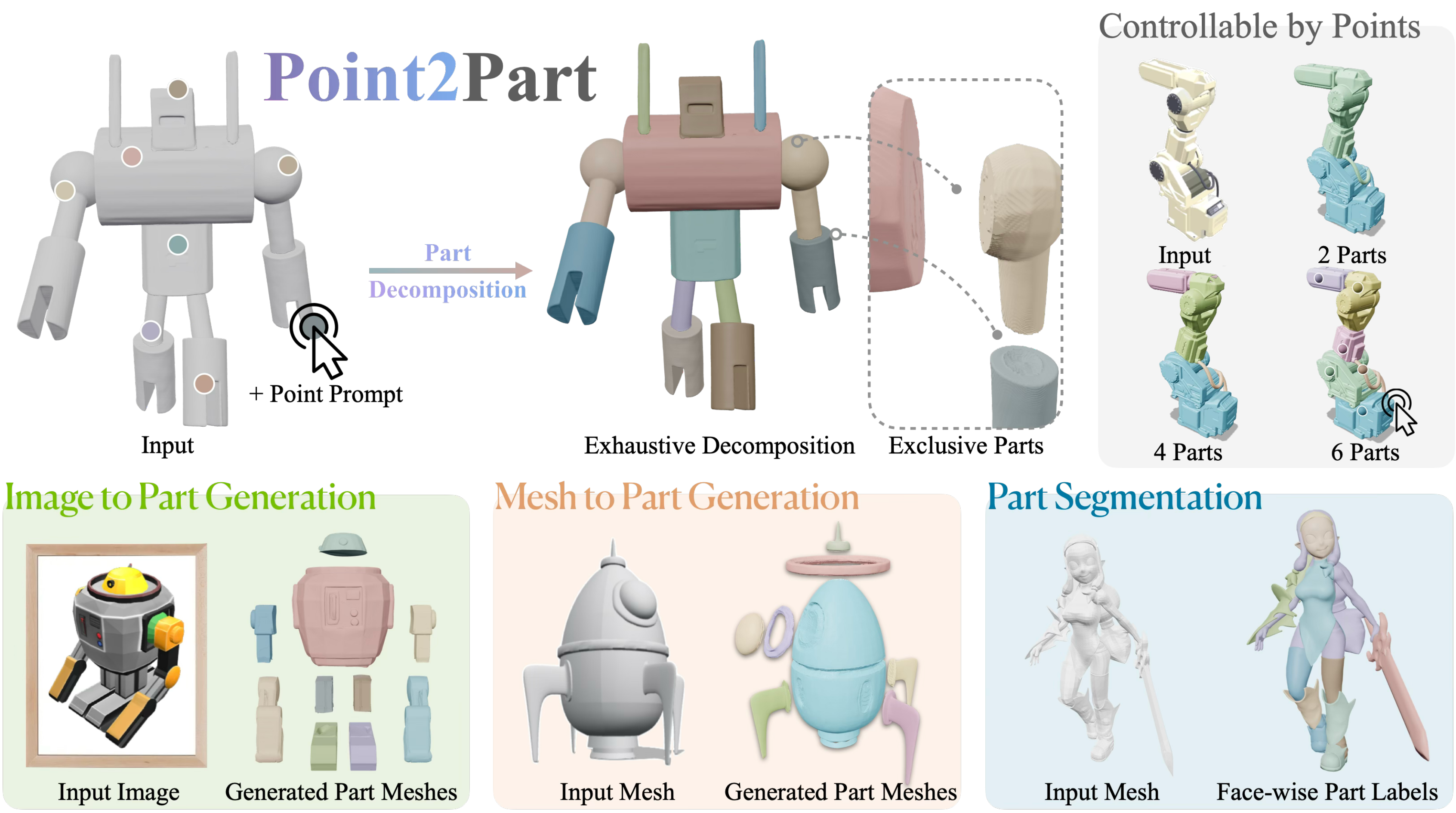}
\vspace{-1.2em}
\caption{\textbf{Unified 3D Partitioning from Point Prompts.}
\textit{(Top)} We support 3D point prompts to specify desired parts and partition the input into closed, exhaustive, and mutually exclusive parts.
\textit{(Right)} Users can interactively control the partition through point prompts.
\textit{(Bottom)} We support image-to-part generation, mesh-to-part generation, and part segmentation within a single model.}
\label{fig:teaser}
\end{figure}

\begin{abstract}
Existing 3D part decomposition methods do not necessarily partition the original shape into non-overlapping parts that collectively cover the entire shape, allowing overlaps or gaps that hinder downstream part-level applications. We instead formulate part decomposition as a \textit{joint partitioning of the entire shape}, in which the predicted parts are non-overlapping and collectively reconstruct the entire shape. Our key insight is that part decomposition should consider all desired parts jointly, rather than modeling each part independently. To this end, we develop a promptable model for 3D part decomposition from images or meshes. Users can specify desired parts through 3D point prompts for controllable decomposition. Given one point prompt per desired part, our model produces the corresponding parts as a complete partition of the entire shape. We build on a pretrained 3D generation model and first obtain a shape latent from either an input image or mesh. We then introduce a prompt encoder that maps each 3D point prompt to a part token while attending to the shape latent. To decode the desired parts, we propose a novel part decoder that jointly scores the entire shape against all part tokens in a coarse-to-fine manner, assigning every position within the shape volume to exactly one part. We perform part decomposition in this shared shape latent space, enabling a unified model for image-to-part generation, mesh-to-part generation, and part segmentation. Our method outperforms existing works on all part-quality metrics across all three tasks, and improves compatibility among parts by an order of magnitude over previous SOTA methods. 
Code is available at \url{https://henrytsui000.github.io/Point2Part}.
\end{abstract}

\newpage
\section{Introduction}
\label{sec:intro}

Recent 3D generative models produce high-quality object-level meshes from a text prompt or a single image~\citep{triposg,trellis2,hunyuan3d21,sam3d}. However, downstream applications such as part-level editing require decomposing these objects into distinct components~\citep{partnet,structurenet}. This has motivated growing interest in part-level 3D generation~\citep{cubepart,omnipart}. Yet existing part-level methods typically produce a collection of parts that may overlap, without jointly considering whether these parts form a disjoint partition of the entire shape.

Existing part generators \citep{partcrafter,partpacker,omnipart,holopart,xpart,cubepart} may use cross-part attention to model relationships between parts, but still decode each part separately, allowing the same spatial region to be occupied by multiple parts (\emph{i.e.,} penetration) or by no part at all (\emph{i.e.,} gaps). They typically control the desired decomposition through a predefined number of parts, text prompts, or 2D image masks~\citep{cubepart,omnipart}, offering indirect or limited control. Other methods support prompt-based 3D part segmentation~\citep{p3sam,pointsam,s2am3d}, which segments the mesh surface into parts based on 3D point prompts placed directly on the surface, but they typically predict a separate face mask independently for each prompt. While this provides direct 3D control over the desired parts, the resulting masks often require post-processing to resolve overlaps and are not constrained to jointly form a complete, non-overlapping segmentation of the object.
Consequently, both approaches may produce overlapping parts or missing components (see \cref{fig:qualitative}), rather than forming a coherent decomposition of the whole object.

To address this, we formulate part decomposition as \textit{a joint partitioning of the entire shape}, where the resulting parts are \emph{exclusive}, meaning their volumes do not overlap, and \emph{exhaustive}, meaning their union recovers the entire volume of the object (see \cref{eq:partition-def} for the formal definition).
The key to enforcing this structure is to define the decomposition jointly over the whole shape, rather than generating each part independently. Given an input image or mesh, users specify each desired part with a 3D point prompt, and we propose a model that decomposes the whole shape into the prompted parts. Our model builds on a pretrained 3D generation model~\citep{hunyuan3d21} and consists of three components: a shape latent backbone, a prompt encoder, and a part decoder. The shape latent backbone represents the input image or mesh as a shape latent capturing the entire geometry, while the prompt encoder maps each point prompt to a corresponding part token while attending to the shape latent. The part decoder then jointly scores the entire shape against all part tokens, assigning every query point within the shape volume to exactly one part. We further refine part boundaries at higher resolutions in a coarse-to-fine manner. This finally yields an exclusive and exhaustive partition by construction.

By representing both image and mesh inputs in a shared shape latent space, our model supports three tasks in a unified way: image-to-part generation, mesh-to-part generation, and part segmentation of mesh surface (see \cref{fig:teaser} \textit{bottom}). For generation, the model generates a closed mesh for each prompted part, while for segmentation, it assigns each mesh face to one of the parts. Moreover, our model offers interactive part decomposition, where users can flexibly control the decomposition by adjusting the number and locations of 3D point prompts (see \cref{fig:teaser} \textit{right}). We evaluate all three tasks on benchmark datasets~\citep{sampart3d}, and our method outperforms state-of-the-art (SOTA) generation and segmentation methods on all part-quality metrics across all three tasks, while improving geometric compatibility between parts by an order of magnitude.

In summary, we make the following contributions:
\begin{enumerate}
\item We formulate 3D part decomposition as a joint partition of the entire shape, identifying the overlap and missing components issues of existing part-level methods, yielding an exclusive and exhaustive partition by construction.
\item We introduce a promptable partition design that represents each user-specified part with a part token. Given a 3D point prompt per part, the model jointly assigns every point in the shape to exactly one prompted part, producing a complete partition of the shape.
\item By leveraging a shared shape latent, we support image-to-part generation, mesh-to-part generation, and part segmentation within a unified framework, achieving SOTA performance and substantially improving geometric compatibility between parts.
\end{enumerate}

\newpage
\section{Related Work}
\label{sec:related}

\paragraph{Promptable part segmentation.}
Promptable segmentation was popularized in 2D by SAM~\citep{sam}, where a click yields a mask, and extended to concept prompts by SAM~3~\citep{sam3}.
In 3D, Point-SAM~\citep{pointsam} brings the click interface to point clouds, P3-SAM~\citep{p3sam} operates natively on meshes and merges per-click masks by suppression and flood fill, PartSAM~\citep{partsam} scales the design with a triplane encoder, and S$^2$AM3D~\citep{s2am3d} adds a continuous scale signal.
Prompt-free methods cluster a learned feature field at a chosen count~\citep{partfield}, lift SAM~2 masks from rendered views onto the mesh~\citep{samesh, sampart3d}, or repurpose a generative model as a segmenter~\citep{segvigen}, and PartObjaverse-Tiny~\citep{sampart3d} is a common benchmark among them.
Each prompt is answered independently, so the masks may overlap or leave gaps, and in 3D, a mask is only a patch on the existing surface, not a closed part.
In contrast, our model answers all prompts jointly as one partition, and it not only segments the surface but also generates each part as a closed mesh.

\paragraph{Part-level generation.}
Different from segmentation, part-level generation produces a closed mesh per part from a given image or mesh, building on whole-shape generators~\citep{triposg,trellis2,hunyuan3d21,sam3d,seed3d2}.
They differ in what specifies the parts.
PartCrafter~\citep{partcrafter} and PartPacker~\citep{partpacker} take only a part count and generate compositional latents, X-Part~\citep{xpart} and FullPart~\citep{fullpart} take a box layout and generate each part within its box, OmniPart~\citep{omnipart} lifts 2D masks from the input image, CubePart~\citep{cubepart} takes a list of part names, and HoloPart~\citep{holopart} and UniPart~\citep{unipart} work from a segmentation, given to the former and generated jointly with the geometry by the latter.
To our knowledge, these methods produce closed geometry, but none constrains one part to stay out of another or reports whether it does, and control is indirect, through counts, boxes, names, or masks.
We decode from a single whole-shape field with user points into exclusive and exhaustive parts.

\paragraph{Primitive-based decompositions.}
An older line represents shapes using cuboids~\citep{tulsiani2017primitives}, superquadrics~\citep{superquadrics}, learned convexes~\citep{cvxnet,bspnet}, or deformed spheres~\citep{neuralparts}.
These methods approximate shapes with a small set of geometric primitives rather than recovering the part decomposition, so their outputs are not directly comparable under our part-level metrics.
\section{Method}
\label{sec:method}

\subsection{Problem Formulation}
\label{sec:method:formulation}

Given a 3D object, represented either by a complete mesh or an image, together with user-specified 3D point prompts, we seek to decompose the object into one 3D part per prompt such that the parts form an exclusive and exhaustive partition of the object.
We first formulate part decomposition as a partition of the object, and then describe how point prompts specify the desired partition.

\paragraph{Part decomposition as a partition.}
We represent the object as a solid $\Omega \subset \mathbb{R}^3$ with surface $\partial\Omega$, and its decomposition as $N$ closed solids $\{\Omega_1,\dots,\Omega_N\}$. For a mesh input, $\Omega$ is the solid enclosed by the input mesh; for an image input, it is the solid enclosed by the mesh generated from the image. Each part is represented as a closed solid, and the complete object is recovered by assembling all parts. 
Let $\operatorname{vol}(\cdot)$ denote the volume of a set.
We formally define the decomposition such that distinct parts have no volume overlap (\textit{exclusive}) and their union recovers the entire object (\textit{exhaustive}):
\begin{equation}
  \underbrace{\operatorname{vol}(\Omega_i \cap \Omega_j) = 0,
  \quad \text{for } i \neq j}_{\text{\emph{exclusive}}},
  \qquad
  \underbrace{\bigcup_{j=1}^{N} \Omega_j = \Omega}
  _{\text{\emph{exhaustive}}}.
  \label{eq:partition-def}
\end{equation}

\paragraph{Point-prompted decomposition.} 
A 3D mesh admits valid decompositions at different granularities. To provide fine-grained and flexible control, we enable users to interactively control the decomposition by directly placing \textit{3D point prompts} on the object $\Omega$, with one prompt for each desired part. In contrast to indirect controls such as text~\citep{cubepart}, 2D image masks~\citep{omnipart}, or part counts~\citep{partcrafter}, 3D point prompts provide finer control over the entire object, including regions that may be occluded in a 2D view. Our model aims to decompose $\Omega$ to the prompted parts to produce a partition satisfying~\cref{eq:partition-def}. For \emph{generation}, each prompt yields a closed part mesh, while for \emph{segmentation}, each face is assigned a part label corresponding to a prompt.

\begin{figure}[t]
\centering
\includegraphics[width=\linewidth]{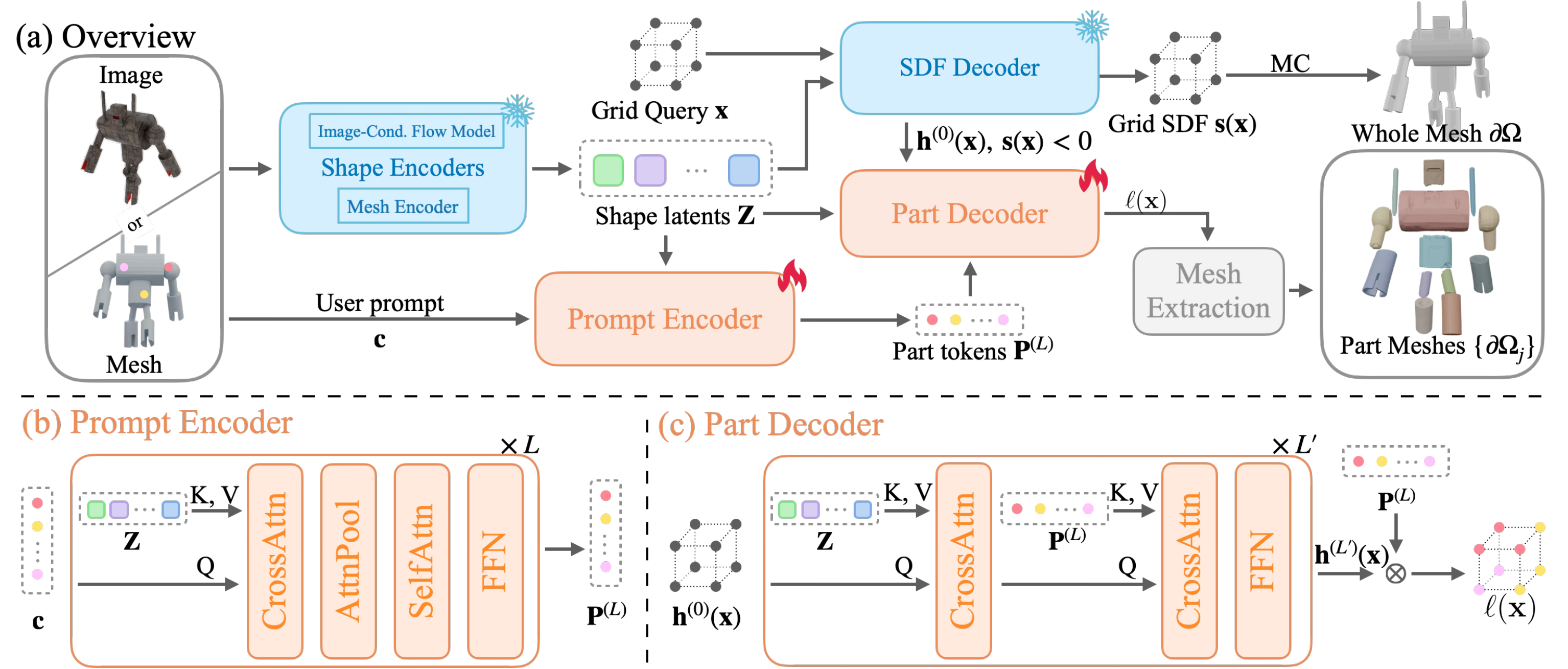}
\caption{\textbf{Method overview.} (a) Given a mesh or an image, a shape encoder first encodes it to shape latents $\mathbf{Z}$; the prompt encoder processes user prompts by attending to shape latents $\mathbf{Z}$ to get part tokens $\mathbf{P}^{(L)}$; the part decoder scores query points $\mathbf{x}$ with the part tokens to predict their part assignments; mesh extraction converts these predictions into exclusive and exhaustive closed part meshes. (b, c) One layer of the prompt encoder and the part decoder, respectively.}
\label{fig:overview}
\end{figure}

\subsection{Model}
\label{sec:method:model}

\paragraph{Overview.}
\Cref{fig:overview} presents our unified model for promptable part generation and segmentation.
Given a mesh or an image together with $N$ user prompts, our model generates either $N$ closed part meshes or assigns one part label to each input face for the \emph{segmentation} task. 
Our model builds on a pretrained latent shape model~\citep{hunyuan3d21} and extends it with a prompt encoder and a part decoder. 
For either input modality (\ie, image or mesh), the backbone provides shape latents $\mathbf{Z}$ in a shared latent space.
For image input, the decoded whole mesh is used to place 3D prompts, while the same shape latents $\mathbf{Z}$ are used for decomposition.
The prompt encoder processes user prompts by attending to $\mathbf{Z}$, producing one token $\mathbf{P}^{(L)}$ per part, and the part decoder jointly scores each query point $\mathbf{x}$ against all part tokens, producing part assignments $\ell(\mathbf{x})$. 
These part assignments $\ell(\mathbf{x})$ derive the exclusive and exhaustive output parts ${\Omega_j}$ by mesh extraction. 

\paragraph{Shape Backbone.} The shape backbone consists of shape encoders and an SDF decoder, as shown in~\cref{fig:overview}. The shape encoder maps the given image or mesh to shape latents, and the SDF decoder reconstructs a full mesh from shape latents. 
For either a mesh or an image input, our shape encoders produce $M$ shape latents
$\mathbf{Z} \in \mathbb{R}^{M \times C}$ in the same latent space, where $C$ is the channel dimension.
For mesh input, $\mathbf{Z}$ is obtained from a mesh encoder~\citep{hunyuan3d21}; for image input, it is predicted by the image-conditioned flow model~\citep{hunyuan3d21}. We leverage pretrained alignment between the shape latents of the two inputs and perform the decomposition on the shared shape latent $\mathbf{Z}$, allowing the same prompt encoder and part decoder to operate on either input.

With the encoded shape latent $\mathbf{Z}$, SDF decoder provides this whole mesh $\partial \Omega$ through its signed distance field, \ie, for a spatial location \(\mathbf{x}\in\mathbb{R}^{3}\), its SDF decoder predicts
\begin{equation}
\mathbf{s}(\mathbf{x})
=
f_0\!\left(
\operatorname{CrossAttn}\!\left(\phi(\mathbf{x}), \mathbf{Z}\right)
\right),
\label{eq:sdf}
\end{equation}
where $\phi$ is a positional embedding, $\phi(\mathbf{x})$ serves as the query, $\mathbf{Z}$
provides the keys and values, and $f_0$ is a linear readout.
The signed distance is negative inside the shape, defining
$\Omega = \{\mathbf{x} \in \mathbb{R}^3 : \mathbf{s}(\mathbf{x}) < 0\}$.
For the subsequent decomposition, we initialize the query feature at each
interior location $\mathbf{x} \in \Omega$ as
$\mathbf{h}^{(0)}(\mathbf{x}) = \phi(\mathbf{x}) \in \mathbb{R}^C$.
A mesh is recovered by evaluating $\mathbf{s}$ on a regular grid and extracting the
zero level set $\partial\Omega$ with marching cubes~\citep{marchingcubes}.

\paragraph{Prompt Encoder.} The prompt encoder takes point prompts~$\mathbf{c}$ for $N$ parts and predicts one part token $\mathbf{P}^{(L)} \in \mathbb{R}^{N \times C}$, corresponding to each specified part by attending to shape latent and inter-part relationships. \Cref{fig:overview}(b) illustrates the prompt encoder.
A point prompt for each part consists of $K$ points for each $j$-th part for a total of $N$ parts. 
Each point is embedded independently as \begin{equation}
  \mathbf{P}^{(0)}
  =
  \big[\phi(\mathbf{c}_{j,k})\big]_{j \le N,\; k \le K}
  \in \mathbb{R}^{N \times K \times C},
  \label{eq:query}
\end{equation}
where $\phi$ is the backbone positional embedding.
A user prompt built from positional embeddings alone carries no information about the shape or about the other prompts.
The prompt encoder supplies both, letting each point prompt reference the asset through $\mathbf{Z}$ and relate to the other prompts.
Its first layer lets every point attend to $\mathbf{Z}$, aggregate the $K$ tokens of each part by attention pooling, and then lets the parts attend to one another. The remaining layers repeat both steps,
\begin{equation}
\begin{aligned}
  \mathbf{P}^{(1)}
  &=
  \mathrm{FFN}\Big(
  \mathrm{SelfAttn}\big(
  \mathrm{Pool}_{k}\,
  \mathrm{CrossAttn}(\mathbf{P}^{(0)}, \mathbf{Z})
  \big)\Big), \\
  \mathbf{P}^{(l+1)}
  &=
  \mathrm{FFN}\Big(
  \mathrm{SelfAttn}\big(
  \mathrm{CrossAttn}(\mathbf{P}^{(l)}, \mathbf{Z})
  \big)\Big),
  \qquad l = 1,\dots,L-1,
\end{aligned}
\label{eq:ground}
\end{equation}
where $\mathrm{Pool}_{k}$ aggregates the $K$ point tokens of each part, and $\mathbf{P}^{(l)} \in \mathbb{R}^{N \times C}$ for $l \ge 1$.
Pooling after cross-attention lets each point gather local geometry before forming one token per part.

\paragraph{Part Decoder.}
Given the part tokens $\mathbf{P}^{(L)}$ from the prompt encoder, shape latents $\mathbf{Z}$, and an interior query point $\mathbf{x}\in\Omega$, the part decoder produces one part assignment per part by relating $\mathbf{x}$ to the local geometry in $\mathbf{Z}$ and to all prompted parts.
These are collected as $\ell(\mathbf{x}) \in \mathbb{R}^N$, indicating how strongly $\mathbf{x}$ belongs to each prompted part, as illustrated in \Cref{fig:overview}(c).
Starting from $\mathbf{h}^{(0)}(\mathbf{x})=\phi(\mathbf{x})$, the decoder refines the query feature through $L'$ layers.
Each layer first attends to $\mathbf{Z}$ to gather local geometry and then to $\mathbf{P}^{(L)}$ to incorporate part-specific information,
\begin{equation}
\mathbf{h}^{(l+1)}(\mathbf{x})
=
\mathrm{FFN}\Big(\mathrm{CrossAttn}\big(
\mathrm{CrossAttn}(\mathbf{h}^{(l)}(\mathbf{x}), \mathbf{Z}),
\mathbf{P}^{(L)}\big)\Big),
\quad l = 0, \ldots, L'-1,
\label{eq:part-decoder}
\end{equation}
where $\mathbf{x} \in \Omega$.
The part assignments are dot products between $\mathbf{h}^{(L')}$ and part tokens $\mathbf{P}^{(L)}$,
\begin{equation}
\ell_j(\mathbf{x})
=
\frac{1}{\sqrt{C}}
\big\langle
f_h(\mathbf{h}^{(L')}(\mathbf{x})),
f_s(\mathbf{P}_j^{(L)})
\big\rangle,
\quad
j = 1, \ldots, N,\quad \mathbf{x} \in \Omega,
\label{eq:logits}
\end{equation}
where $f_h$ and $f_s$ are learned projections, and each part token acts as a dynamic linear classifier for its part, as in Mask2Former~\citep{mask2former} but on interior points.
One pass over the queries decodes all $N$ parts, so the cost is set by the number of queries and grows only slightly with $N$.

\paragraph{Mesh Extraction.}
Before decoding $\ell(\mathbf{x})$ into part meshes, note that although the backbone of~\cref{eq:sdf} regresses a signed distance, and mesh extraction uses only its zero level set, and hence only the sign of $\mathbf{s}$, a binary decision at every point, inside or outside.
Extracting parts therefore needs one further decision on the interior $\Omega$, and \cref{eq:partition-def} fixes its form: exclusive and exhaustive together say that every point of $\Omega$ is assigned to exactly one part, that is, a single-valued map $\pi: \Omega \to \{1,\dots,N\}$ defined on all of $\Omega$. 
Any such map satisfies~\cref{eq:partition-def}, since
\begin{equation}
  \sum_{j=1}^{N} \mathbf{1}[\pi(\mathbf{x}) = j] = 1 \quad \forall\, \mathbf{x} \in \Omega,
  \label{eq:onehot}
\end{equation}
which implies the exclusivity and exhaustiveness conditions in \cref{eq:partition-def}.
We take $\pi$ to be the argmax of the part assignments of~\cref{eq:logits}, the part with the largest score at each point,
\begin{equation}
  \pi(\mathbf{x}) = \argmax_{j \in \{1,\dots,N\}} \ell_j(\mathbf{x}),
  \qquad
  \Omega_j = \{\mathbf{x} \in \Omega : \pi(\mathbf{x}) = j\}.
  \label{eq:partition}
\end{equation}
Since \cref{eq:partition-def} follows from the form of $\pi$ and not from its values, it holds whatever the model predicts, without using suppression, flood filling, or an overlap penalty.
The part meshes follow from the same field.
Marching cubes~\citep{marchingcubes} extracts the zero level set of a field, the surface between its negative and non-negative regions, so a closed mesh of part $j$ needs a field that is negative exactly on $\Omega_j$.
We obtain it from the whole field by keeping $\mathbf{s}$ where $\pi = j$ and discarding its sign elsewhere,
\begin{equation}
  \mathbf{s}_j(\mathbf{x}) =
  \begin{cases}
    \mathbf{s}(\mathbf{x}) & \pi(\mathbf{x}) = j, \\
    |\mathbf{s}(\mathbf{x})| & \text{otherwise},
  \end{cases}
  \label{eq:partfield}
\end{equation}
so that $\mathbf{s}_j < 0$ exactly on $\Omega_j$, since outside the object $\mathbf{s}$ is already non-negative and inside the other parts the sign is dropped.
Evaluating $\mathbf{s}_j$ on the grid that covers $\Omega$ and running marching cubes returns one closed mesh per part, whose faces on $\partial\Omega$ coincide with the whole and whose faces where $\Omega_j$ meets another part are the cut surfaces that close it.

\label{sec:method:training}
\paragraph{Supervision and sampling.}
We train the model to assign each location inside the whole to a single GT part.
To support a flexible number of points per prompt, during training we sample $K \sim \mathcal{U}\{1,\dots,4\}$ surface points $\{\mathbf{c}_{j,k}\}_{k=1}^{K}$ for each part prompt from the $j$-th GT part.
Unless otherwise specified, we use $K=1$ at inference.
Given the signed distance $d_j(\mathbf{x})$ to each annotated part, we define the target assignment as
\begin{equation}
  y(\mathbf{x}) = \argmin_{j \in \{1,\dots,N\}} d_j(\mathbf{x}), \qquad \mathbf{x} \in \Omega .
  \label{eq:label}
\end{equation}
This assigns a unique label to interior locations, resolving overlaps in the annotations. We concentrate training queries near part boundaries, where the assignment is most ambiguous, while also sampling the shape interior and regions on and near its surface. 

\paragraph{Training objective.}
We train the model to predict the prompted part assignment for each interior query, while encouraging part tokens from the same part to be consistent and query features to be discriminative.
We optimize
\begin{equation}
  \mathcal{L} = \mathcal{L}_{\mathrm{cls}} + \lambda_{\mathrm{dice}}\,\mathcal{L}_{\mathrm{dice}} + \lambda_{\mathrm{cons}}\,\mathcal{L}_{\mathrm{cons}},
  \label{eq:total}
\end{equation}
where $\mathcal{L}_{\mathrm{cls}}$ and $\mathcal{L}_{\mathrm{dice}}$ supervise the predicted partition, and $\mathcal{L}_{\mathrm{cons}}$ encourages consistency across prompts for the same part.
Since each interior location is assigned to the part with the highest score, we formulate $\mathcal{L}_{\mathrm{cls}}$ as an $N$-way classification loss.
Let $p_j(\mathbf{x}) = \mathrm{softmax}_j\,\ell(\mathbf{x})$ denote the predicted probability of assigning $\mathbf{x}$ to part $j$.
Given the target assignments $y(\mathbf{x})$, we apply a focal cross-entropy~\citep{focal} over the labeled training queries $\mathcal{X} \subset \Omega$,
\begin{equation}
  \mathcal{L}_{\mathrm{cls}} = -\frac{1}{|\mathcal{X}|}\sum_{\mathbf{x} \in \mathcal{X}} w_{y(\mathbf{x})}\,\big(1 - p_{y(\mathbf{x})}(\mathbf{x})\big)^{\gamma}\,\log p_{y(\mathbf{x})}(\mathbf{x}),
  \label{eq:focal}
\end{equation}
where $w_j$ balances parts of different sizes and $\gamma$ reduces the contribution of easy queries, emphasizing locations whose assignment remains uncertain.
To complement pointwise classification with part-level region supervision, we add a soft Dice loss~\citep{vnet},
\begin{equation}
  \mathcal{L}_{\mathrm{dice}} = \frac{1}{N}\sum_{j=1}^{N} \Bigg(1 - \frac{2\sum_{\mathbf{x} \in \mathcal{X}} p_j(\mathbf{x})\,\mathbf{1}[y(\mathbf{x}) = j] + 1}{\sum_{\mathbf{x} \in \mathcal{X}} p_j(\mathbf{x}) + \sum_{\mathbf{x} \in \mathcal{X}} \mathbf{1}[y(\mathbf{x}) = j] + 1}\Bigg).
  \label{eq:dice}
\end{equation}
A part token should represent the intended part rather than the particular surface locations used to specify it.
We therefore encode two independently sampled prompt sets for the same ground-truth parts, obtaining tokens $\mathbf{P}^{(L)}$ and $\mathbf{P}'^{(L)}$, and encourage their agreement,
\begin{equation}
  \mathcal{L}_{\mathrm{cons}} = \frac{1}{N}\sum_{j=1}^{N} \big\| \mathbf{P}^{(L)}_j - \mathbf{P}'^{(L)}_j \big\|_2^2 .
  \label{eq:cons}
\end{equation}

\subsection{Inference}
\label{sec:method:inference}

\paragraph{Generation task.} 
We support both mesh and image inputs for part generation.
For either input, the predicted partition is converted into $N$ closed part meshes using~\cref{eq:partfield} followed by per-part marching cubes.
For image input, prompts are placed on the generated mesh rather than on the image, so that parts on the back of the object, which the image does not show, can also be assigned.

Following octree-based isosurface extraction~\citep{meagher1982,wilhelms1992} and its adaptation to neural shape decoding in FlashVDM~\citep{flashvdm}, we perform coarse-to-fine refinement of both the whole-shape and part SDFs.
The whole-shape SDF is refined within a narrow surface band, defined by $|s(\mathbf{x})| < \eta$ for a chosen threshold $\eta$.
The part SDFs are refined only where $s(\mathbf{x}) < \eta$ and the two highest part-assignment scores, $\ell_{(1)}(\mathbf{x})$ and $\ell_{(2)}(\mathbf{x})$, differ by less than a margin $\delta$:
\begin{equation}
  \mathcal{R}^{\mathrm{whole}} = \{\mathbf{x} : |s(\mathbf{x})| < \eta\},
  \qquad
  \mathcal{R}^{\mathrm{part}} = \{\mathbf{x} : s(\mathbf{x}) < \eta \ \text{and}\ \ell_{(1)}(\mathbf{x}) - \ell_{(2)}(\mathbf{x}) < \delta\}.
  \label{eq:refine}
\end{equation}

\newpage

Elsewhere, the fine grid inherits the coarse value.
The $N$ part meshes are cut from the one field, and their interiors $\Omega_1,\dots,\Omega_N$ reassemble $\Omega$, a whole from which none of its parts is absent.

\paragraph{Segmentation task.}
We support promptable part segmentation for mesh input.
Given a mesh and one point prompt per part, we query the decoder at each face centroid $\mathbf{x}_f$ instead of $\mathbf{x} \in \Omega$ and assign the face the predicted part label $\pi(\mathbf{x}_f)$, leaving the input mesh geometry unchanged.
\section{Experiments}
\label{sec:experiments}

\subsection{Setup}
\label{sec:experiments:setup}

\paragraph{Baselines.}
We compare against SOTA methods for part generation~\citep{cubepart,holopart,xpart,partcrafter,partpacker,omnipart} and part segmentation~\citep{segvigen,pointsam,partsam,s2am3d,p3sam,partfield}.
To keep the comparison as consistent as possible, we run all baselines using their official checkpoints and inference settings.
As these methods differ in inputs and training data, with some relying on private training data~\citep{partfield}, our quantitative results serve as a proof-of-concept comparison under their respective released settings.
Generation methods~\citep{partcrafter,partpacker} requiring a part count receive the GT part count;
CubePart~\citep{cubepart} and OmniPart~\citep{omnipart} receive part names and 2D segmentation masks, respectively.
Unless stated otherwise, promptable methods~\citep{s2am3d} receive one point per GT part (\ie, $K{=}1$), sampled uniformly from the corresponding part surface where it does not overlap with other parts. For image input, the point is sampled on the generated mesh surface after aligning the generated mesh to the GT mesh. Please refer to~\cref{sec:appendix:metrics} for detailed baseline settings.

\paragraph{Datasets.}
We train on the public dataset HY3D-Bench~\citep{hy3dbench}, which has $240$K assets with part-level labels.
The baselines were trained on their own datasets of comparable or larger size, almost all of which are not released.
We evaluate on PartObjaverse-Tiny~\citep{sampart3d}, a widely used benchmark for 3D part decomposition, with $200$ meshes and face-level part-instance labels, and we remove any assets from the HY3D-Bench training set that overlap with PartObjaverse-Tiny. 
We use the same assets to evaluate all three tasks: a mesh for segmentation and mesh-conditioned generation, and a single rendered view for image-conditioned generation.

\paragraph{Metrics.}
We evaluate three aspects of the decomposition: part quality, whole-shape geometry, and part compatibility.
For part quality, \emph{pCD} and \emph{pF1@$\tau$} measure matched-part Chamfer distance and surface F1; for segmentation, we additionally report face-level \emph{mIoU}, where face correspondence is available.
For whole-shape geometry, \emph{CD} and \emph{F1@.05} compare the union of generated parts with the ground-truth whole.
For part compatibility, we report inter-part penetration (\emph{pen\%}) and the fraction of closed parts (\emph{wt\%}); \emph{pen\%} measures violations of exclusivity, while \emph{pCD} and \emph{pF1} also reflect failures of exhaustiveness (see \cref{sec:appendix:metrics}). 
Distances are reported at $10^{-2}$ scale in the normalized unit cube and averaged over assets.
\emph{Inference time} is measured per asset on one A100, and image predictions are ICP-aligned before evaluation.

\paragraph{Implementation details.}
We train the HY3D-2.1 model~\citep{hunyuan3d21}, and also provide an ablation on using TripoSG~\citep{triposg} as backbone, which has similar architecture to HY3D-2.1;
Please refer to \cref{sec:appendix:impl} for detailed architectures, sampling budgets, loss, \etc.

\subsection{Main Results}
\label{sec:experiments:main}

\begin{table}[t]
\centering
\caption{\textbf{Comparison of our method with SOTA methods for 3D part generation from a mesh (top) and an image (bottom).}
We evaluate part decomposition quality, part compatibility, whole-shape geometry, and inference time.}
\vspace{4pt}
\label{tab:gen}
\small
\setlength{\tabcolsep}{4pt}
\resizebox{\linewidth}{!}{%
\begin{tabular}{lcccccccr}
\toprule
& \multicolumn{3}{c}{Part quality} & \multicolumn{2}{c}{Part compatibility} & \multicolumn{2}{c}{Whole geometry} & Inference \\
\cmidrule(lr){2-4} \cmidrule(lr){5-6} \cmidrule(lr){7-8}
Method & pCD$\downarrow$ & pF1@.01$\uparrow$ & pF1@.05$\uparrow$
       & pen\%$\downarrow$ & wt\%$\uparrow$
       & CD$\downarrow$ & F1@.05$\uparrow$
       & time (s)$\downarrow$ \\
\midrule
\multicolumn{9}{l}{\textit{Mesh input}} \\
CubePart~\citep{cubepart}       & 4.71 & 51.5 & \second{72.5} & 2.09 & \best{100.0} & \second{1.21} & \best{97.5} & \second{29.1} \\
HoloPart~\citep{holopart}       & 5.29 & 46.6 & 73.9 & \second{1.01} & 32.6 & 1.54 & 95.0 & 99.1 \\
X-Part~\citep{xpart}            & \second{4.53} & \second{52.5} & 73.4 & 3.05 & \second{90.8} & \best{1.19} & \second{97.3} & 147.2 \\
\textbf{Ours}                  & \best{2.73} & \best{57.0} & \best{84.4} & \best{0.06} & \best{100.0} & 1.56 & 94.1 & \best{24.4} \\
\midrule
\multicolumn{9}{l}{\textit{Image input}} \\
PartCrafter~\citep{partcrafter} & 12.75 & 6.3 & 29.6 & 2.45 & 70.5 & 3.76 & 76.9 & 84.8 \\
PartPacker~\citep{partpacker}   & 8.11 & 17.6 & 54.2 & 1.25 & 86.0 & 2.22 & \second{91.7} & \best{26.7} \\
OmniPart~\citep{omnipart}       & \second{6.43} & \second{23.9} & \second{60.8} & \second{0.96} & \second{94.7} & \second{2.20} & 91.4 & 73.3 \\
\textbf{Ours}                  & \best{5.34} & \best{27.2} & \best{68.6} & \best{0.01} & \best{100.0} & \best{2.00} & \best{93.3} & \second{37.8} \\
\bottomrule
\end{tabular}}
\end{table}

\begin{table}[t]
\centering
\caption{\textbf{Comparison of our method with SOTA methods for 3D part segmentation.}
We evaluate part decomposition quality on face mIoU and on part-level CD and F1, and inference time.}
\label{tab:seg}
\vspace{4pt}
\small
\begin{tabular}{lccccr}
\toprule
& \multicolumn{4}{c}{Part quality} & Inference \\
\cmidrule(lr){2-5}
Method & mIoU$\uparrow$ & pCD$\downarrow$ & pF1@.01$\uparrow$ & pF1@.05$\uparrow$
       & time (s)$\downarrow$ \\
\midrule
SegviGen~\citep{segvigen}                     & 20.50 & 10.05 & 22.8 & 40.3 & 317.9 \\
Point-SAM~\citep{pointsam}       & 39.12 & 5.37  & 47.7 & 68.7 & 1.2 \\
PartSAM~\citep{partsam}                       & 50.31 & 3.24  & \second{63.4} & 79.2 & 36.6 \\
S$^2$AM3D~\citep{s2am3d}          & 50.43 & \second{3.16} & 62.3 & \second{80.5} & \best{0.3} \\
P3-SAM~\citep{p3sam}                          & 54.66 & 4.33  & 61.0 & 74.8 & 19.1 \\
PartField~\citep{partfield}                   & \second{69.10} & 5.31 & 62.3 & 72.5 & \second{1.1} \\
\midrule
\textbf{Ours}                                          & \best{69.80} & \best{2.10} & \best{78.3} & \best{87.0} & \best{0.3} \\
\bottomrule
\end{tabular}
\end{table}

\begin{figure}[t]
\centering
\includegraphics[width=\linewidth]{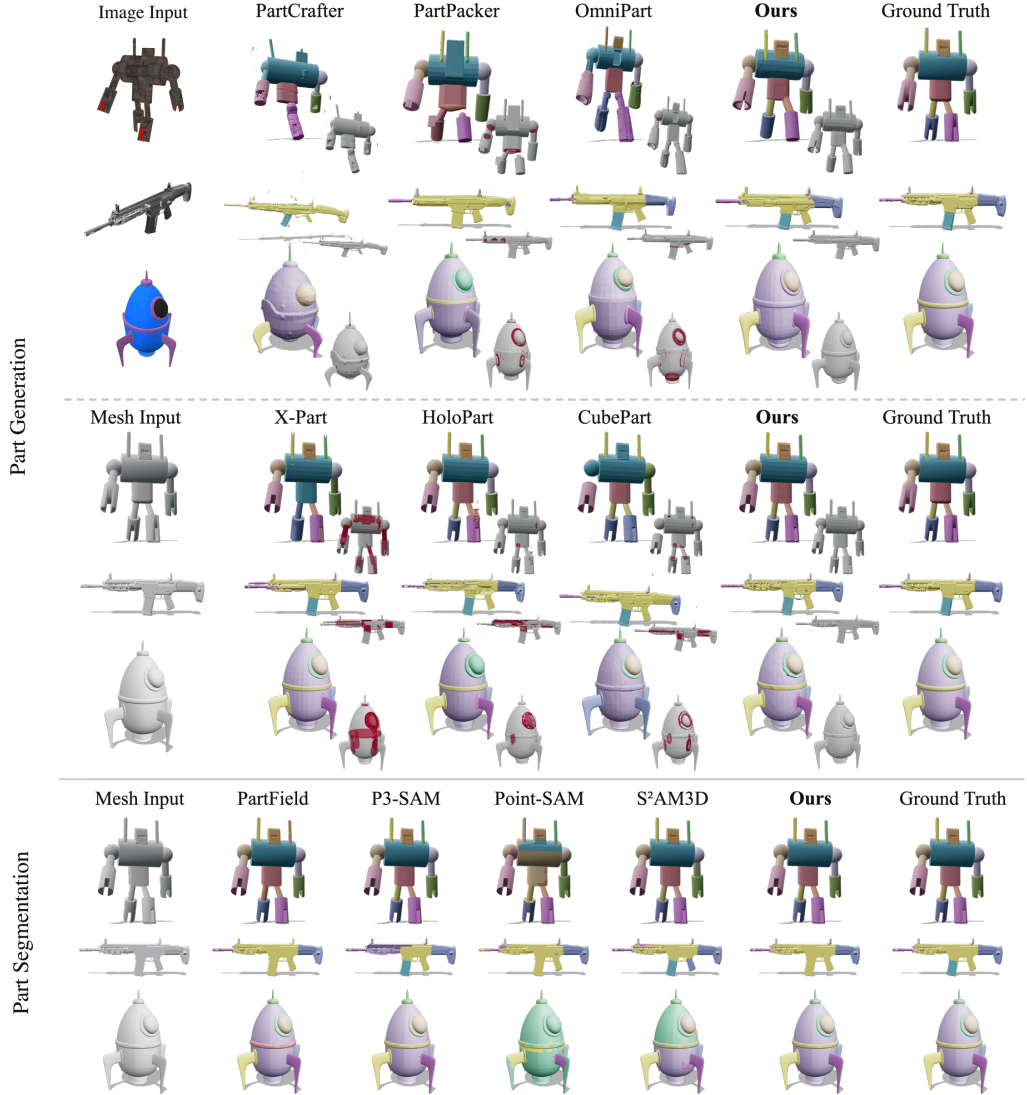}
\vspace{-4pt}
\caption{\textbf{Qualitative comparison of our method with SOTA methods on the three tasks.}
Part generation from an image (top) and from a mesh (middle), and part segmentation (bottom).
The grey mesh beside each generated result marks in red the volume claimed by more than one part.
Prior methods overlap, drop, split, or merge parts, while ours matches the ground truth.}
\label{fig:qualitative}
\end{figure}

\paragraph{Part generation.}
\Cref{tab:gen} compares our method with prior methods for 3D part generation from mesh (top) and image (bottom) inputs.
Our method achieves the best part quality for both inputs and reduces inter-part penetration by over an order of magnitude, demonstrating that our decomposition results are exclusive. 
The small residual penetration comes from marching-cubes interpolation at part seams (\cref{sec:appendix:metrics}). Whole-shape fidelity is bounded by the backbone’s reconstruction of the input.
Our method is the fastest from a mesh and second fastest from an image, since all parts are decoded jointly in one pass.
\Cref{fig:qualitative} visualizes the same behavior, further comparisons in \cref{sec:appendix:vis}.

\paragraph{Part segmentation.}
\Cref{tab:seg} shows that our model leads on every metric of the segmentation task, with the largest gain in pF1 at the tight threshold, meaning our boundaries fall close to the ground-truth boundaries rather than merely in the right region.
It is also the fastest, tied with S$^2$AM3D.
\Cref{fig:qualitative} shows the same on individual assets, and \cref{sec:appendix:vis} provides further comparisons.

\subsection{Ablations}
\label{sec:experiments:ablation}
\begin{table}[!tb]
\centering
\begin{minipage}[t]{0.53\linewidth}
\centering
\caption{\textbf{Model ablations.}
We evaluate part generation from a mesh with the part-quality metrics.}
\label{tab:ablation-components}
\vspace{6pt}
\resizebox{\linewidth}{!}{\footnotesize\setlength{\tabcolsep}{4pt}
\begin{tabular}{lcccc}
\toprule
Ablation & pCD$\downarrow$ & pF1@.01$\uparrow$ & pF1@.05$\uparrow$ & pen\%$\downarrow$  \\
\midrule
(a) w/o prompt enc.                 & 4.24 & 50.4 & 75.2 & 0.07 \\
(b) w/o $\mathcal{L}_{\mathrm{cons}}$  & 3.11 & 53.1 & 80.8 & 0.06 \\
(c) w/o refine.                     & 2.86 & 54.6 & 83.9 & \textbf{0.05} \\
(d) Indep. assign.                     & 2.90 & 54.9 & 83.2 & 1.23 \\
\midrule
\textbf{Ours}                              & \textbf{2.73} & \textbf{57.0} & \textbf{84.4} & 0.06 \\
\bottomrule
\end{tabular}}
\end{minipage}
\hspace{0.01\linewidth}
\begin{minipage}[t]{0.43\linewidth}
\centering
\caption{\textbf{Ablation on the number of prompt points $K$.}
We evaluate part segmentation with the metrics of \cref{tab:seg}.}
\label{tab:ablation-prompts}
\vspace{10pt}
{\scriptsize\setlength{\tabcolsep}{2.2pt}
\begin{tabular}{llcccc}
\toprule
Backbone & $K$ & mIoU$\uparrow$ & pCD$\downarrow$ & pF1@.01$\uparrow$ & pF1@.05$\uparrow$ \\
\midrule
\multirow{2}{*}{TripoSG} & 1   & 67.40 & 2.41 & 75.6 & 85.3 \\
        & 4   & 72.68 & 1.53 & 81.6 & 90.3 \\
\midrule
\multirow{2}{*}{HY3D-2.1} & 1   & 69.80 & 2.10 & 78.3 & 87.0 \\
        & 4   & \textbf{74.84} & \textbf{1.35} & \textbf{84.0} & \textbf{91.8} \\
\bottomrule
\end{tabular}}
\end{minipage}\hfill
\end{table}

\paragraph{Effect of Modules.}
\Cref{tab:ablation-components} evaluates each component of our model.
(a) Removing the prompt encoder causes the largest drop in part quality, showing the importance of letting prompts attend to the shape and other parts.
(b) Removing $\mathcal{L}_{\mathrm{cons}}$ produces a smaller but consistent drop, as prompts on the same part may yield different boundaries.
(c) Removing coarse-to-fine refinement mainly affects tight F1, indicating that refinement corrects small boundary and seam errors.
(d) Independent assignment replaces the joint argmax in \cref{eq:partition} with independent binary assignments from the per-part logits in \cref{eq:logits}, causing substantial inter-part penetration despite similar part quality.

\paragraph{Number of prompt points.}
\Cref{tab:ablation-prompts} shows the effect of the number of points per part across two backbones. Across both backbones, every metric improves as more points are added.
A single point indicates where a part is, while additional points reveal its extent.
\Cref{sec:appendix:results} reports the full sweep.

\paragraph{Limitations.}
Our method faces challenges when decomposing extremely thin structures or large open surfaces (Appendix \Cref{fig:failed}), which may not enclose a well-defined volume and are therefore difficult to represent as solid parts. This limitation is partly due to the pretrained backbone, which struggles to represent and reconstruct such thin and open geometry. Improving the backbone's representation of these geometries could extend our method to a broader range of shapes.

\section{Conclusion}
\label{sec:conclusion}

We introduce a unified promptable model for 3D part generation and segmentation that formulates part decomposition as a partition of the whole rather than a collection of independently generated parts.
By jointly assigning the whole among prompted parts, our model produces an exclusive and exhaustive partition by construction.
A shape can admit multiple valid decompositions; point prompts let users directly specify and control the desired decomposition.
A single model supports generation from meshes or images, as well as mesh segmentation, achieving state-of-the-art part quality while reducing part geometric incompatibility by over an order of magnitude.
\newpage

\bibliography{refs}
\bibliographystyle{iclr2027_conference}

\newpage
\appendix
\crefalias{section}{appendix}
\crefname{appendix}{Appendix}{Appendices}
\Crefname{appendix}{Appendix}{Appendices}

\let\plainsection\section
\renewcommand{\section}{\FloatBarrier\plainsection}

In this appendix, we provide the architecture, sampling, and optimization details in \cref{sec:appendix:impl}, the evaluation protocol including the interpenetration and watertightness tests in \cref{sec:appendix:metrics}, additional results on the number of parts, the number of prompt points, inference time, and a second benchmark in \cref{sec:appendix:results}, and additional qualitative comparisons, exploded views, control of the decomposition, and failure cases in \cref{sec:appendix:vis}.

\section{Implementation and Training Details}
\label{sec:appendix:impl}

\paragraph{Architecture.}
\cref{tab:hyperparams} lists every hyperparameter of the model reported in the main text.
The cross-attention to $\mathbf{Z}$ in each decoder layer starts from the backbone's readout; every other added weight is trained from scratch.
Three of them start at zero so that training begins from the backbone's own behavior: the output projection of each added residual branch, so the whole-shape field is unchanged at initialization; the projection $f_s$, so all $N$ logits are equal; and the attention-pooling query in the first encoder layer, so pooling begins as the mean of the $K$ tokens.

\begin{table}[h]
\centering
\caption{\textbf{Hyperparameters of our model.}
Symbols follow the notation of \cref{sec:method}, and -- marks a value with no symbol in the text.}
\label{tab:hyperparams}
\small
\begin{tabular}{llcr}
\toprule
& Hyperparameter & Symbol & Value \\
\midrule
\multirow{6}{*}{Architecture}
& Shape Latents (TripoSG / Hunyuan3D) & $M$ & 2048 / 4096 \\
& Channels & $C$ & 1024 \\
& Surface samples per shape & -- & 20{,}480 \\
& Prompt encoder layers & $L$ & 2 \\
& Part decoder layers & $L'$ & 3 \\
& Attention heads & -- & 8 \\
\midrule
\multirow{12}{*}{Training}
& Points per prompt & $K$ & $\mathcal{U}\{1,\dots,4\}$ \\
& Queries per asset & -- & 5376 \\
& \quad surface / near surface / bounding box & -- & 1024 / 1024 / 512 \\
& \quad part surface / inside / part boundary & -- & 768 / 1024 / 1024 \\
& Focal exponent & $\gamma$ & 2 \\
& Loss weights & $\lambda_{\mathrm{dice}}$, $\lambda_{\mathrm{cons}}$ & 1, 0.5 \\
& Batch size, epochs & -- & 256, 10 \\
& Learning rate (base / readout) & -- & $2.4 \times 10^{-4}$ / $4 \times 10^{-4}$ \\
& Schedule (warm-up steps, decay fraction) & -- & 150, 0.2 \\
& EMA decay & -- & 0.999 \\
\midrule
\multirow{4}{*}{Inference}
& Coarse / fine grid & -- & $128^3$ / $512^3$ \\
& narrow-band & $\eta$ & 0.05 \\
& margin & $\delta$ & 0.2 \\
& Points per prompt & $K$ & 1 \\
\bottomrule
\end{tabular}
\end{table}

\paragraph{Query sampling.}
Training samples its queries instead of evaluating the field on a grid, since most grid cells lie far from any surface or seam, where the loss is already zero.
We draw the queries of each asset from the pools of \cref{sec:method:training}, which allocate most of the budget to the surface and the seams between parts, while keeping a sparse, uniform pool for the rest of the bounding box.

\begin{figure}[ht]
\centering
\includegraphics[width=\linewidth]{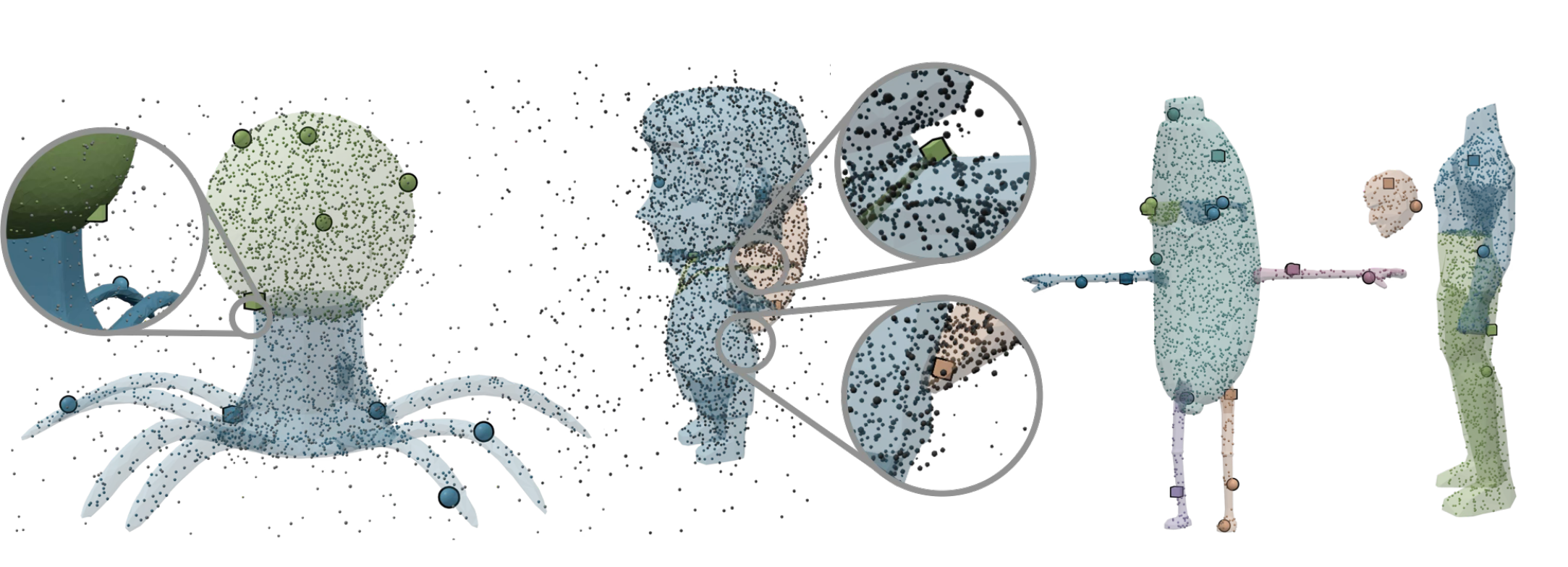}
\caption{\textbf{Training query sampling and prompt points.}
Small points are the sampled queries, colored by part; the larger markers are the prompt points, with balls and cubes marking two prompt sets for the same parts.}
\label{fig:query-pools}
\end{figure}

\Cref{fig:query-pools} shows the results for four assets; the leftmost panel shows the queries crowding the seams.
The prompt points, one to four per part, are drawn on each part's surface where they do not overlap any other part.
Two prompt sets are drawn per asset, the balls and the cubes in the figure, and the consistency loss $\mathcal{L}_{\mathrm{cons}}$ links their partitions.

\paragraph{Training implementation details.}
Training runs on four A100 80GB GPUs for 3 days.
The learning rate follows a warm-up, hold, and decay schedule, rising over the first steps, staying constant for most of training, and decaying over the final fifth.
We keep an exponential moving average of the trained weights and use the averaged weights at inference, which removes the step-to-step variation of the raw weights without changing the model.
The base rate and the averaging constant are listed in \cref{tab:hyperparams}.

\section{Metric Details}
\label{sec:appendix:metrics}

\paragraph{Baselines.}
Every baseline runs its released checkpoint with its default settings.
The one exception is PartField: it evaluates 20 cluster counts per asset and selects the one with the highest mIoU against the ground truth. We instead use the annotated number of parts as a fixed count.
FullPart~\citep{fullpart}, and UniPart~\citep{unipart} are discussed in \cref{sec:related} but not compared, since their inference code has not been fully released.

\paragraph{Interpenetration.}
We sample $2{,}000$ points on the surface of every generated part with a fixed seed.
A probe is kept only if it is still inside its own part under the fast-winding-number sign~\citep{fwn}; a probe that passed through a thin part tells nothing about that part, and without this guard $99.98$ percent of the flagged probes on a test set were of this kind.
Writing $Q_j$ for the surviving probes of part $j$ and $\Omega_k$ for the solid bounded by part $k$, a probe counts as penetrating if it lies inside any other part by the same test, which involves no grid or resolution, and
\begin{equation}
  \mathrm{pen} = \frac{\sum_{j} \#\{\mathbf{q} \in Q_j : \exists\, k \neq j,\ \mathbf{q} \in \Omega_k\}}{\sum_{j} |Q_j|},
  \label{eq:pen}
\end{equation}
in percent, per object and averaged over the benchmark.
Our parts partition one volume, so the residual we report is marching-cubes interpolation across a seam within one cell, whereas methods that decode each part as an independent field report interpenetration proper; for methods that output open shells the winding number is fractional and the value is approximate.

\paragraph{Watertightness.}
\emph{wt\%} is the share of generated parts whose mesh is closed, every edge shared by exactly two faces, per object and averaged over the benchmark.
A part that is not closed has no interior, so it cannot be moved or replaced as a solid, and a decomposition whose parts are not all closed does not partition the whole.

\paragraph{Alignment for image input.}
A shape generated from an image lives in the generator's own frame, so image-input methods are aligned to the ground truth before scoring, while mesh-input methods use the identity.
Both meshes are normalized by their bounding box, centered and scaled so that the longest side is one, and rigid ICP~\citep{icp} on $10$K surface points per side is run from $24$ axis-aligned initial poses, without scale or reflection, keeping the transform of lowest cost.
The same transform is applied to the whole, to every part, and to the part-mesh vertices; the ground truth is only normalized.

\paragraph{Matching and face labels.}
Generated parts are matched to ground-truth parts one to one by the Hungarian algorithm on the part Chamfer distance, and pCD and pF1 are computed on the matched pairs.
For mIoU, each face of the ground-truth mesh takes the label of the generated part that projects onto it, which gives a face labelling for generated parts and for predicted labels alike.

\section{Additional Results}
\label{sec:appendix:results}

\begin{figure}[h]
\centering
\includegraphics[width=0.49\linewidth]{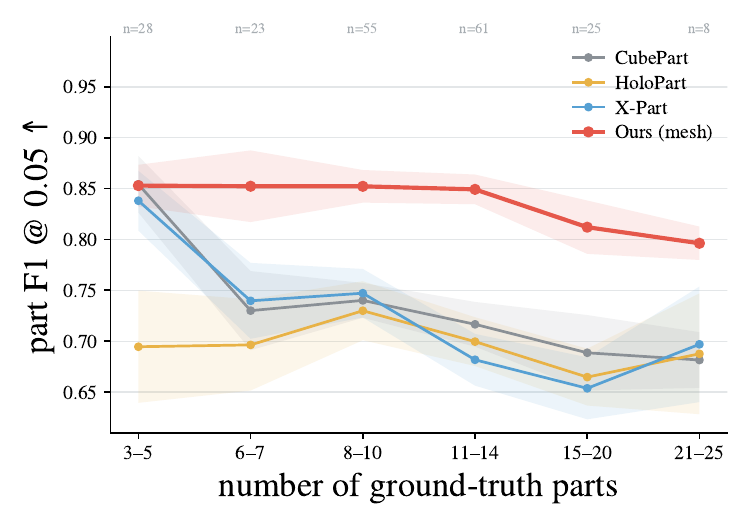}\hfill
\includegraphics[width=0.49\linewidth]{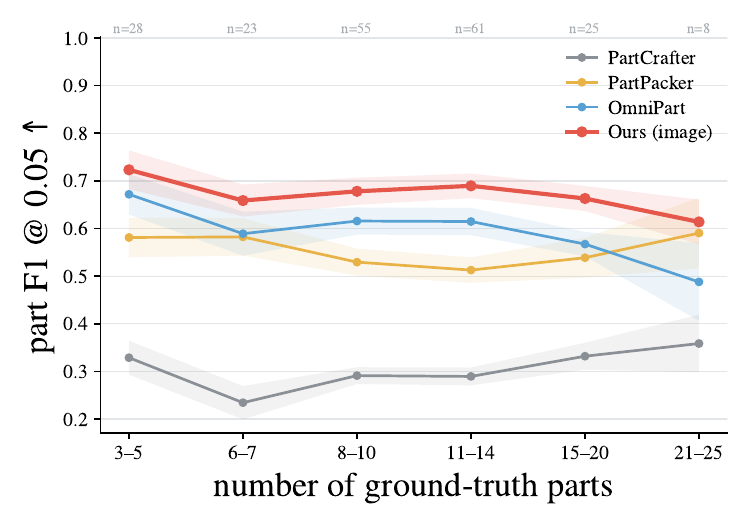}\\[2pt]
\includegraphics[width=0.49\linewidth]{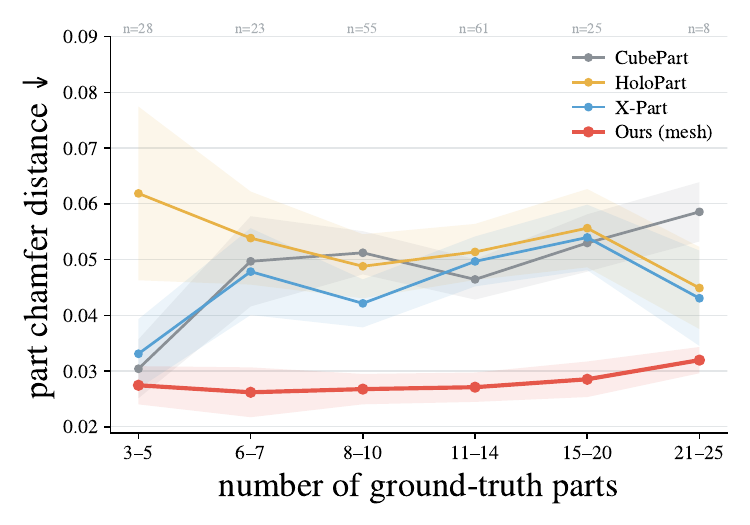}\hfill
\includegraphics[width=0.49\linewidth]{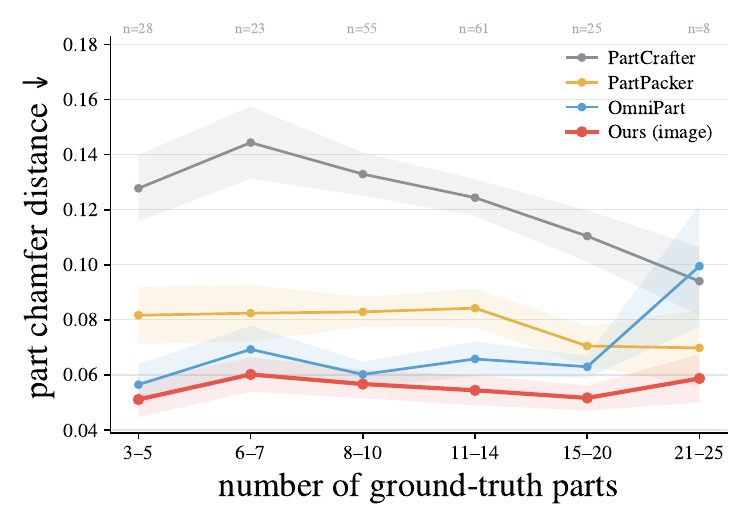}
\caption{Part quality against the number of ground-truth parts, from a mesh (left) and from an image (right), with part quality metrics.}
\label{fig:parts-vs-n}
\end{figure}

\paragraph{Number of parts.}
\Cref{fig:parts-vs-n} bins the benchmark assets by their number of ground-truth parts and plots part quality per bin.
From a mesh, every generator loses accuracy as the part count grows, whereas ours stays close to its level on the simplest objects, so the margin is smallest on objects with a handful of parts and largest on those with more than twenty.
From an image, the ordering is the same, and OmniPart, the closest baseline on few parts, falls below PartPacker on the objects with the most parts.
A partition does not become harder to keep exclusive and exhaustive as $N$ grows, since every point is still assigned once, and the figure shows that accuracy follows.

\begin{figure}[ht]
\centering
\includegraphics[width=0.49\linewidth]{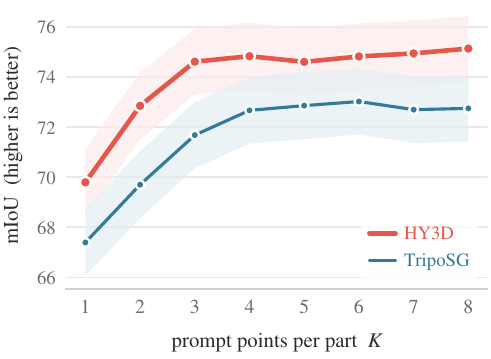}\hfill
\includegraphics[width=0.49\linewidth]{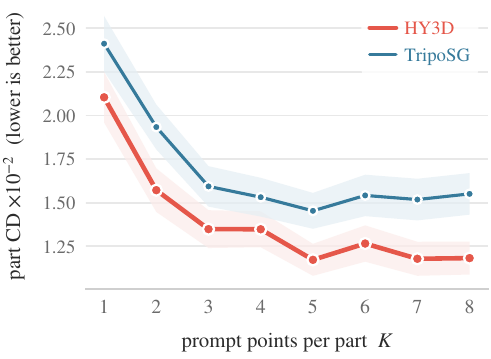}
\caption{Segmentation quality against the number of prompt points per part $K$, for both backbones, with mIoU (left) and pCD (right).
Both metrics improve steeply up to three points and change little beyond.}
\label{fig:k-sweep}
\end{figure}

\paragraph{Number of prompt points.}
\Cref{fig:k-sweep} extends \cref{tab:ablation-prompts} to $K$ from one to eight points per part.
Both metrics improve steeply from one to three points and then level off, and the gap between the backbones stays constant across $K$, so extra points mainly remove the ambiguity of a single click rather than add information the backbone lacks.
Since the decoder cost is independent of $K$, the measured end-to-end runtime changes negligibly.

\begin{figure}[ht]
\centering
\includegraphics[width=0.49\linewidth]{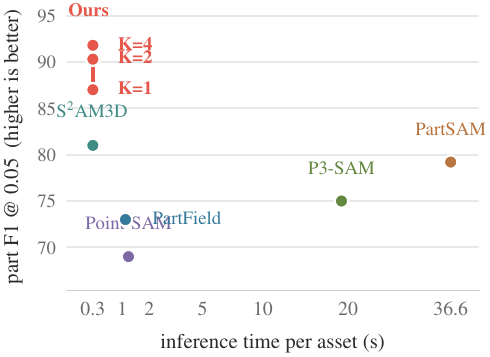}\hfill
\includegraphics[width=0.49\linewidth]{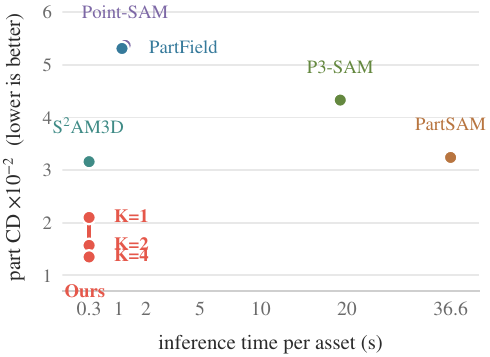}
\caption{Segmentation quality against inference time per asset, pF1@.05 (left) and pCD (right), with our model at $K \in \{1, 2, 4\}$.
Ours is the most accurate at close to the lowest time, and raising $K$ improves quality at no cost in time.}
\label{fig:latency}
\end{figure}

\paragraph{Inference time.}
\Cref{fig:latency} plots segmentation quality against inference time per asset.
Our model is the most accurate and under a second per asset.
The three points for $K \in \{1, 2, 4\}$ lie on a vertical line, since the points of a prompt are pooled into one token in the first encoder layer and the decoder cost does not depend on $K$.

\paragraph{Other benchmarks.}
\Cref{tab:partnext} repeats the segmentation comparison on the PartNeXt evaluation set~\citep{partnext}, a benchmark with finer and hierarchical part labels from a different source than PartObjaverse-Tiny.
Ours still leads every column among the methods, indicating that the gains transfer to a dataset from a different source.

\begin{table}[ht]
\centering
\caption{\textbf{Comparison of our method with SOTA methods for 3D part segmentation on PartNeXt.}
We evaluate part decomposition quality on face mIoU and on part-level pCD and pF1.}
\label{tab:partnext}
\begin{tabular}{lcccc}
\toprule
Method & mIoU$\uparrow$ & pCD$\downarrow$ & pF1@.01$\uparrow$ & pF1@.05$\uparrow$ \\
\midrule
Point-SAM~\citep{pointsam} & 44.41 & 4.24 & 60.8 & 78.8 \\
PartField~\citep{partfield}  & 52.55 & 7.25 & 53.0 & 65.7 \\
S$^2$AM3D~\citep{s2am3d}& 52.99 & \second{2.66} & 69.9 & \second{85.1} \\
P3-SAM~\citep{p3sam} & \second{54.89} & 3.24 & \second{72.6} & 83.1 \\
\midrule
\textbf{Ours} & \best{57.76} & \best{2.39} & \best{75.7} & \best{86.1} \\
\bottomrule
\end{tabular}
\end{table}

\paragraph{Reproducibility.} We randomly sampled point prompts and repeated the experiment over five independent runs. The resulting standard deviations are small: 0.08 for pCD, 0.4 for both pF1@.01 and pF1@.05, and 0.3 for mIoU in the segmentation task. This indicates that the reported results are stable with respect to prompt sampling.

\section{Additional Visualization}
\label{sec:appendix:vis}

\begin{figure}[h]
\centering
\includegraphics[width=\linewidth]{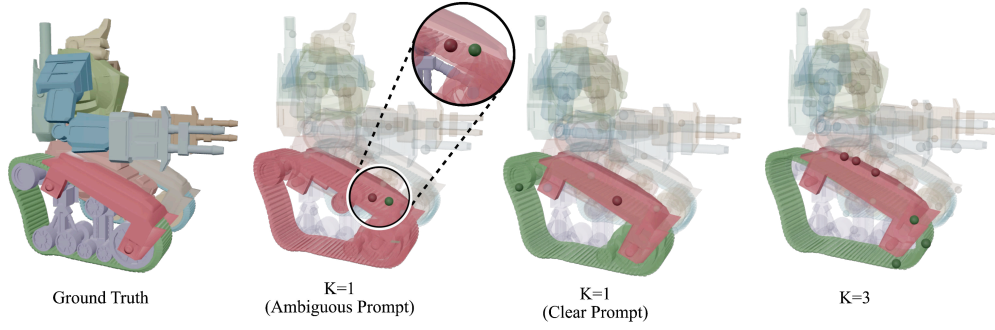}
\caption{\textbf{Failure case and the effect of prompt placement.}
The same mesh is decomposed from one point per part placed near a contact between two parts (second column), from one point per part placed on the body of each part (third column), and from three points per part (fourth column).}
\label{fig:limitation}
\end{figure}

\begin{figure}[h]
\centering
\includegraphics[width=\linewidth]{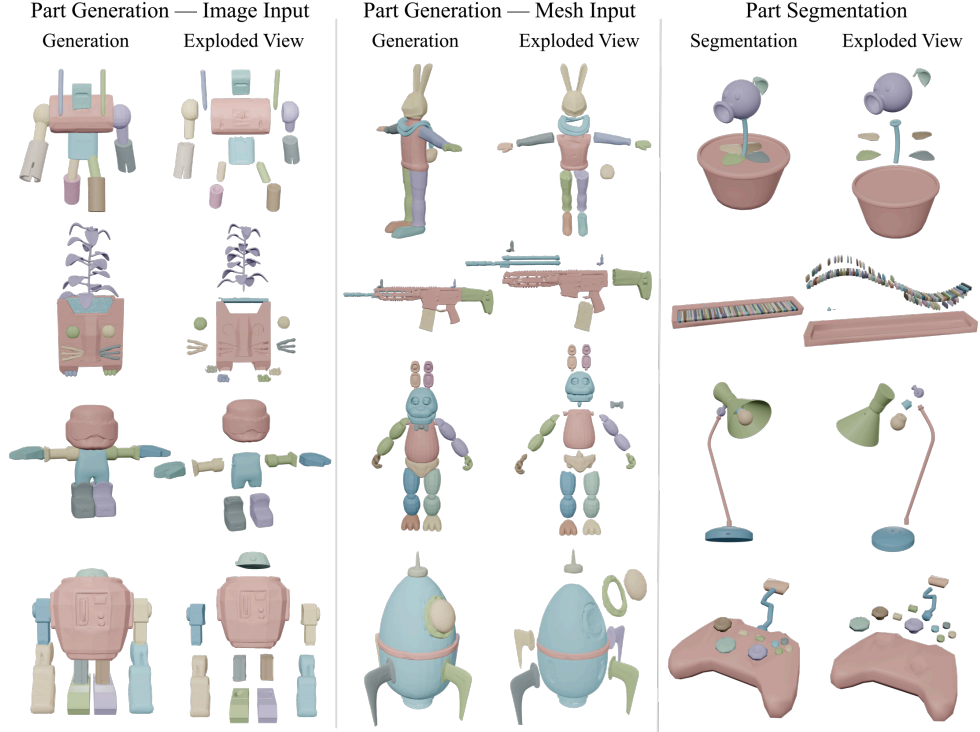}
\caption{\textbf{Exploded views of our results.}
Three assets per task, shown assembled and with the parts moved apart: part generation from an image (left), from a mesh (middle), and part segmentation of a mesh (right).}
\label{fig:exploded}
\end{figure}

\paragraph{Exploded views.}
\Cref{fig:exploded} moves the parts of our results apart.
For generation, from an image or a mesh, every part is a closed mesh with its own cut faces, so the parts separate without any repair and reassemble into the whole, which is the use the introduction asks for.
For segmentation, the parts are the labeled faces of the input mesh, so they are open surfaces, and the exploded view shows where the label boundaries fall.

\begin{figure}[ht]
\centering
\includegraphics[width=\linewidth,height=0.9\textheight,keepaspectratio]{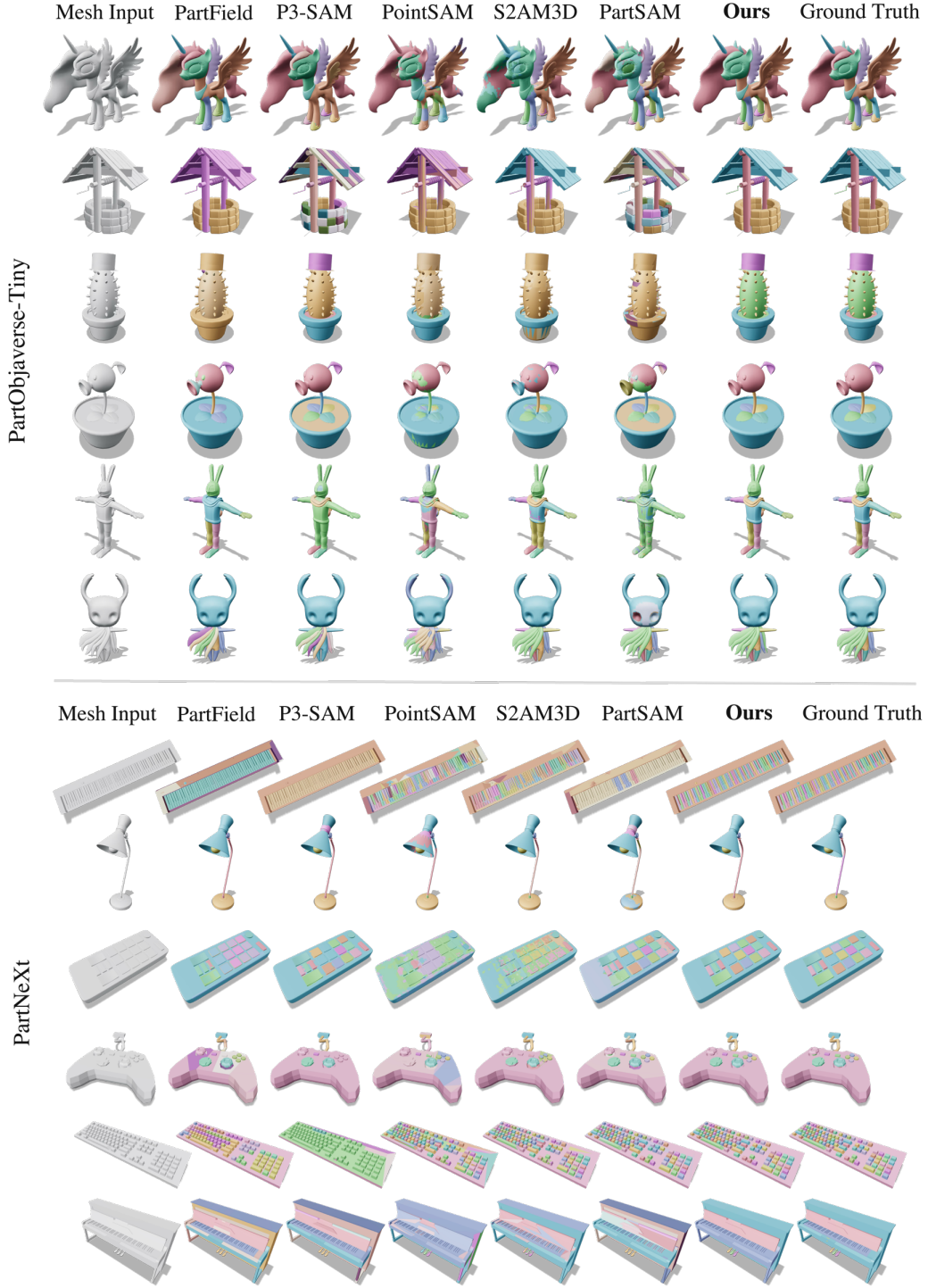}
\caption{\textbf{Part segmentation on PartObjaverse-Tiny (top) and PartNeXt (bottom)}. One point prompt per part, against the segmenters in \cref{tab:seg} and the ground truth.
Ours keeps small repeated parts separate, such as the spikes, keys, and buttons, where the baselines merge or fragment them.}
\label{fig:qual-seg}
\end{figure}

\paragraph{Segmentation.}
\Cref{fig:qual-seg} compares per-face labels on assets from both segmentation benchmarks.
The baselines err in two ways: merging small repeated parts into their neighbor, as with the cactus spikes and the keyboard keys, or fragmenting a single part into several, as with the roof and the piano body, while a partition with one prompt per part does neither.

\begin{figure}[ht]
\centering
\includegraphics[width=\linewidth,height=0.9\textheight,keepaspectratio]{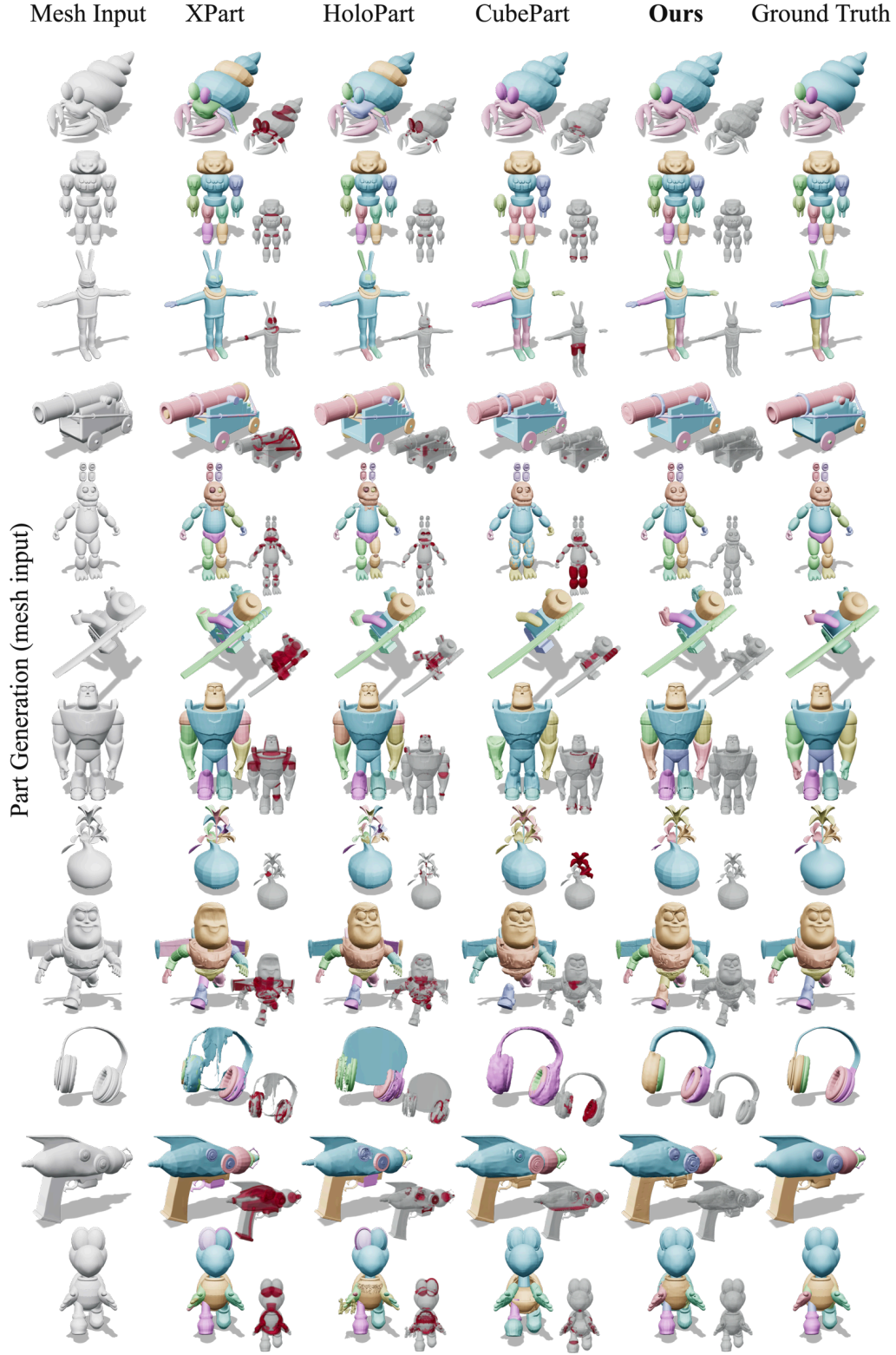}
\caption{\textbf{Part generation from a mesh on further assets}. Compared against the generators in \cref{tab:gen}, with the volume claimed by more than one part marked in red beside each result.
Every baseline interpenetrates at the joints, whereas our parts meet with negligible overlap and follow the ground-truth decomposition.}
\label{fig:qual-meshgen}
\end{figure}

\begin{figure}[ht]
\centering
\includegraphics[width=\linewidth,height=0.9\textheight,keepaspectratio]{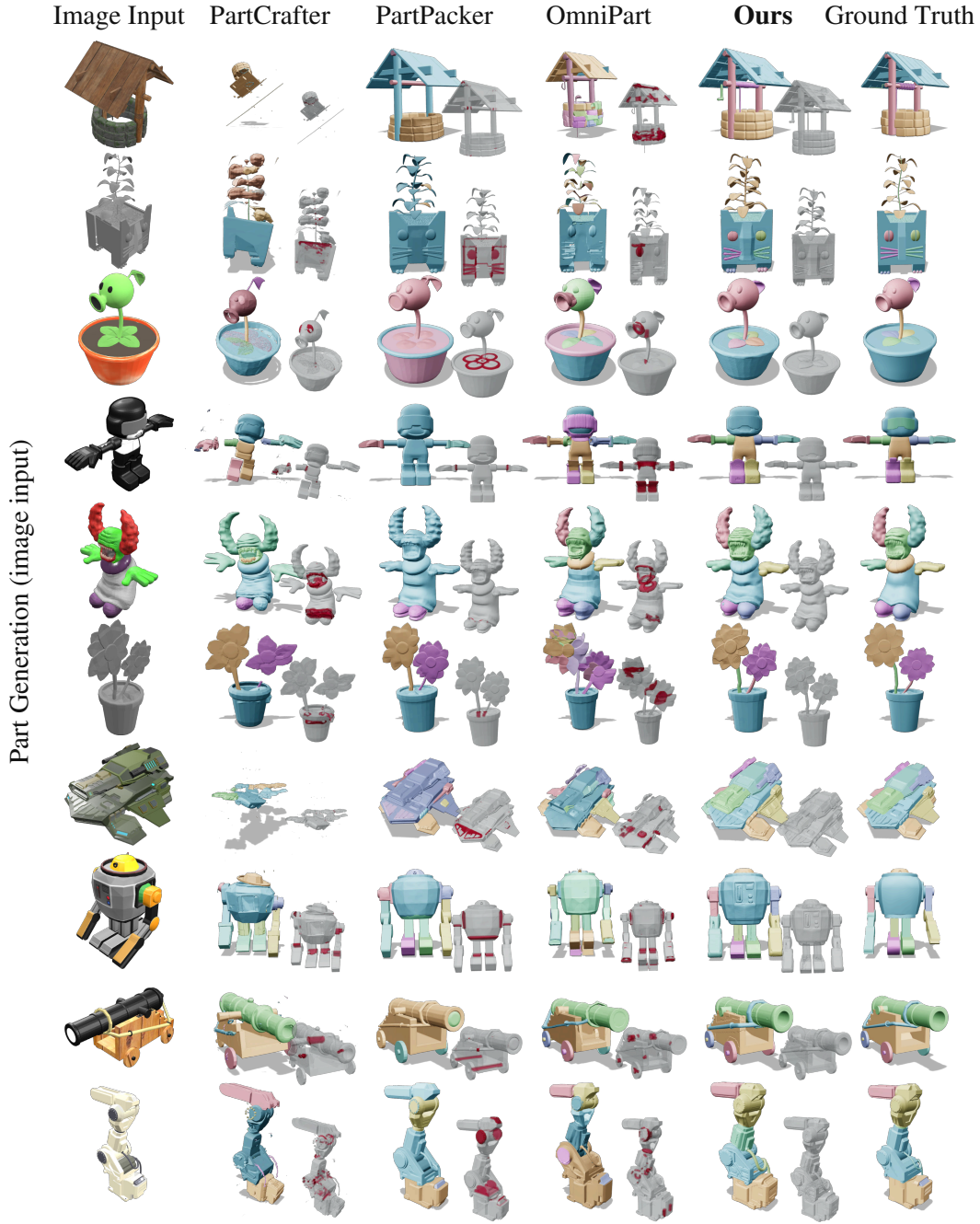}
\caption{\textbf{Part generation from a single image on further assets}. Compared against the generators in \cref{tab:gen}, with the volume claimed by more than one part marked in red beside each result.
The baselines interpenetrate at the joints or return a fragmentary object, whereas our parts meet with negligible overlap and follow the ground-truth decomposition.}
\label{fig:qual-imggen}
\end{figure}

\paragraph{Generation.}
\Cref{fig:qual-meshgen} extends \cref{fig:qualitative} to twelve further assets from a mesh, and \cref{fig:qual-imggen} to ten assets from an image.
The red regions, the volume claimed by two parts, appear at the joints of every baseline and are absent from ours, and where a baseline drops a part or returns a fragmentary object, ours keeps the decomposition of the ground truth.

\begin{figure}[ht]
\centering
\includegraphics[width=\linewidth]{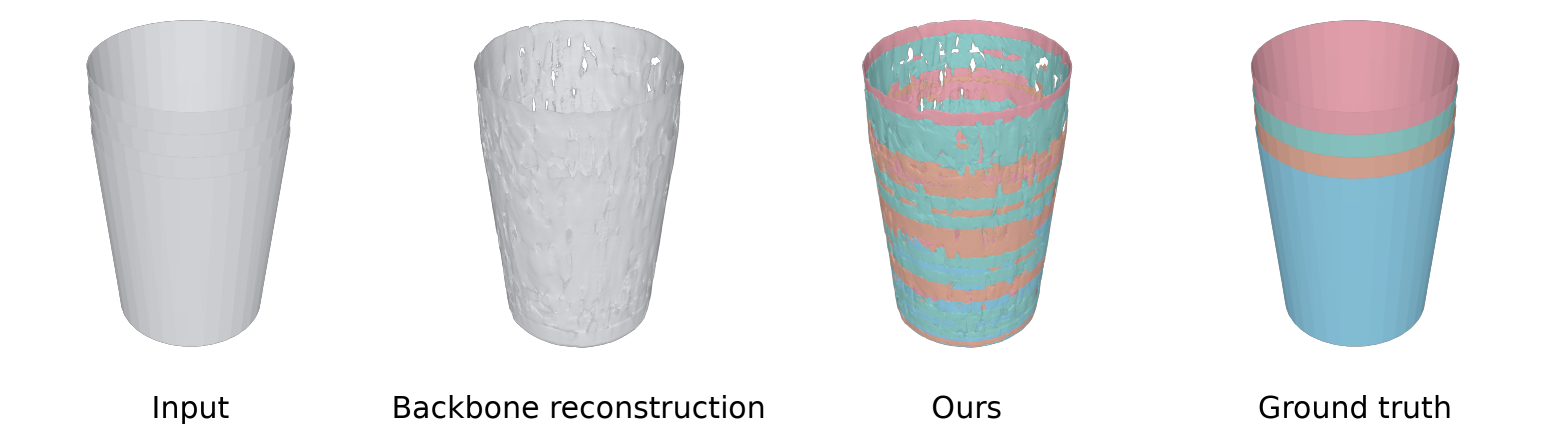}
\caption{\textbf{Limitation on thin and open geometry.}
Large open surfaces and thin shells are poorly reconstructed by the pretrained VAE, and our decomposition consequently inherits these geometric errors.
From left to right: input, VAE reconstruction, our decomposition, and ground truth.}
\label{fig:failed}
\end{figure}

\begin{figure}[ht]
\centering
\includegraphics[width=\linewidth,height=0.9\textheight,keepaspectratio]{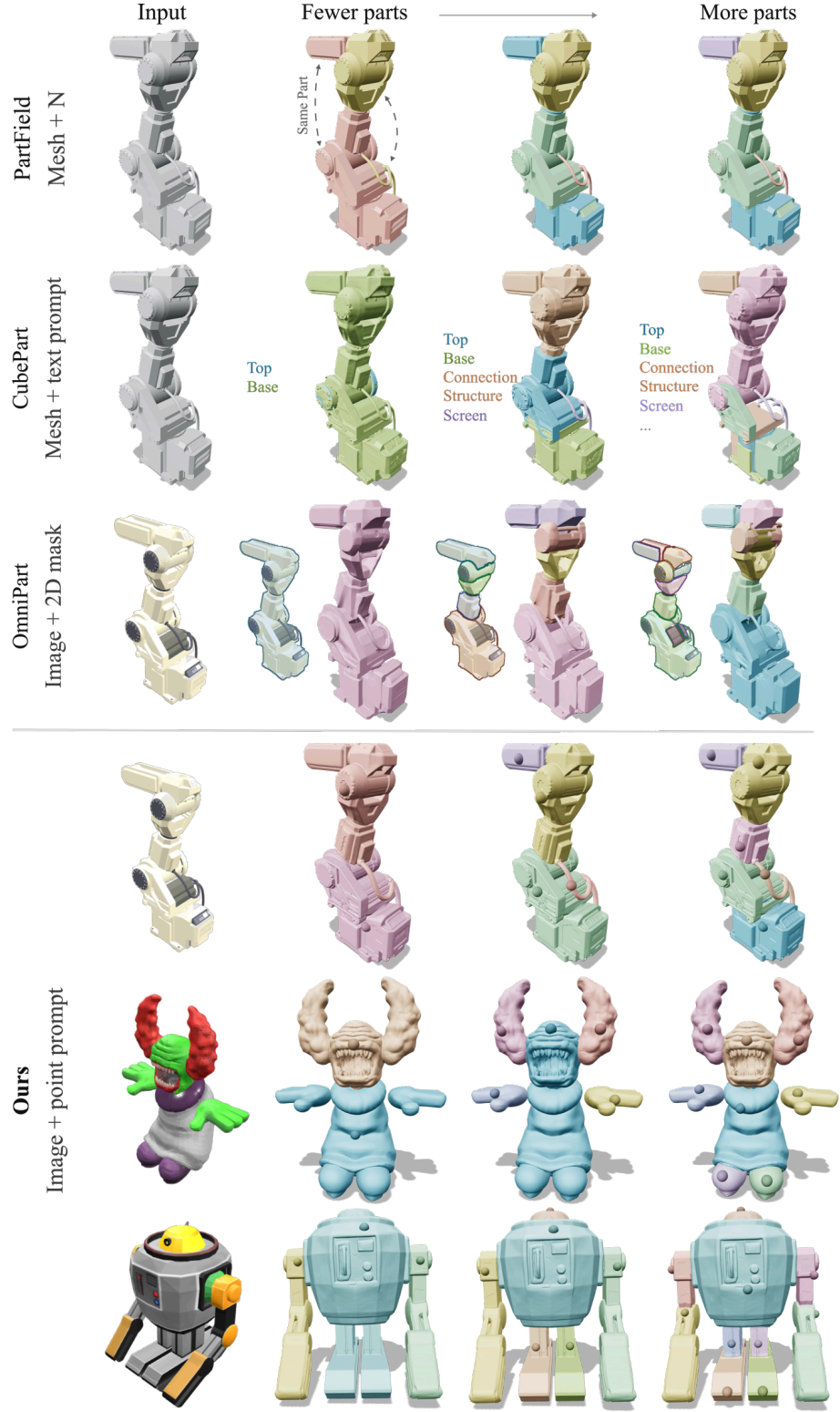}
\caption{\textbf{Controlling the decomposition.}
Objects are decomposed at three granularities, from fewer parts to more parts, through the specification each method takes: a part count for PartField, part names for CubePart, 2D masks on the image for OmniPart, and one point per part for ours.
The top shows the same sample under every method; the bottom shows ours on two further samples.}
\label{fig:control}
\end{figure}

\paragraph{Controlling the decomposition.}
\Cref{fig:control} decomposes objects at three granularities, from fewer to more parts, through the specification each method takes.
The top block shows one object, a robot arm, under every method.
A part count fixes how many parts PartField returns but not which, so the boundaries at each count are chosen by its clustering rather than by the user.
Part names pass through CubePart's reading of them, and the number of parts it returns does not follow the number of names.
Masks for OmniPart are drawn on the image, so only parts visible in the view can be specified.
With one point per part, each added point splits off the part it sits on and leaves the other parts in place, so the user moves from a coarse to a fine decomposition by adding points.
The bottom block shows the same on two further objects, from the image alone.

\paragraph{Failure cases and prompt placement.} \Cref{fig:limitation} shows the failure mode we observe most often.
An ambiguous prompt, a point placed where two parts touch, gives both prompts the same neighborhood, and the decoder merges the parts (second column).
A clear prompt, a point on the body of each part, avoids this (third column), and more points per prompt recover from an ambiguous one, since the pooled prompt then covers the part rather than one location (fourth column, as in \cref{tab:ablation-prompts}).

\end{document}